\documentclass[11pt,a4paper]{article}
\usepackage[hyperref]{eacl2021}
\usepackage{times}
\usepackage{latexsym}

\usepackage{graphicx}
\usepackage{epstopdf}
\usepackage{hyperref}
\usepackage[utf8]{inputenc}
\usepackage[arabic,USenglish]{babel}

\usepackage{microtype}
\usepackage{color}
\usepackage{cleveref}

\definecolor{pinky}{rgb}{0.87, 0.19, 0.39}

\aclfinalcopy 
\def\aclpaperid{18} 

\title{Let-Mi: An Arabic Levantine Twitter Dataset for Misogynistic Language}

\author{Hala Mulki \\
  The ORSAM Center\\
 for Middle Eastern Studies\\
 Ankara, Turkey \\
  \texttt{hala.mulki@orsam.org.tr} \\\And
  Bilal Ghanem \\
  University of Alberta\\
  Edmonton, Canada \\
                               \\
  \texttt{bilalhgm@gmail.com} \\}

\date{}

\begin{document}
\maketitle

\begin{abstract}
Online misogyny has become an increasing worry for Arab women who experience gender-based online abuse on a daily basis.\ Misogyny automatic detection systems can assist in the prohibition of anti-women Arabic toxic content.\ Developing such systems is hindered by the lack of the Arabic misogyny benchmark datasets.\ In this paper, we introduce an Arabic Levantine Twitter dataset for Misogynistic language (LeT-Mi) to be the first benchmark dataset for Arabic misogyny.\ We further provide a detailed review of the dataset creation and annotation phases.\ The consistency of the annotations for the proposed dataset was emphasized through inter-rater agreement evaluation measures.\ Moreover, Let-Mi was used as an evaluation dataset through binary/multi-/target classification tasks conducted by several state-of-the-art machine learning systems along with Multi-Task Learning (MTL) configuration.\ The obtained results indicated that the performances achieved by the used systems are consistent with state-of-the-art results for languages other than Arabic, while employing MTL improved the performance of the misogyny/target classification tasks.
\end{abstract}

\section{Introduction}
\label{Intro}
Social media are the digital tribunes where people can express their thoughts and opinions with the ultimate freedom.\ However, this freedom comes at a price, especially when it enables spreading abusive language and hate speech against individuals or groups.\ Misogyny is one type of hate speech that disparages a person or a group having the female gender identity; it is typically defined as hatred of or contempt for women~\citep{Nockleby,moloney2018}.\ According to \cite{bailey2016}, based on the misogynistic behavior, misogynistic language can be classified into several categories such as discredit, dominance, derailing, sexual harassment, stereotyping and objectification, and threat of violence.

During the last decade, online misogynistic language has been recognized as a universal phenomenon spread across social media platforms such as Facebook and Twitter.\ Similar to their peers worldwide, women in the Arab region are subjected to several types of online misogyny, through which gender inequality, lower social status, sexual abuse, violence, mistreatment and underestimation are, unfortunately, reinforced and justified.\ Moreover, in specific contexts, online misogyny may evolve into systematic bullying campaigns, launched on social media to attack and, sometimes, violently threaten women who have a powerful influence over the society; as it is the case with female journalists/reporters ~\citep{ferrier2018}.

The automatic detection of online misogynistic language can facilitate the prohibition of anti-women toxic contents.\ While many efforts have been spent in this domain for Indo-European languages, the development of Arabic misogyny detection systems is hindered by the lack of the Arabic annotated misogyny resources.\ Building such resources involves several challenges in terms of data collection and annotation, especially with the drastic lexical variations among the Arabic dialects and the ambiguity introduced by some underrepresented dialects such as the Levantine dialect.\ Levantine is one of the five major varieties of Arabic dialects. People in Syria, Lebanon, Palestine, and Jordan are considered the native speakers of Levantine with minor differences from one country to another~\citep{sleiman2018}.

The 17\textsuperscript{th} October 2019 protests in Lebanon witnessed a heavy use of Twitter especially by the journalists \cite{proteststats}.\ While covering the protests on the ground, many female journalists were, unfortunately, prone to gender-based abuse receiving misogynistic replies for the tweets they post.




In this paper, we introduce the first Arabic \textbf{Le}vantine \textbf{T}witter dataset for \textbf{Mi}sogynistic language (LeT-Mi) to be a benchmark dataset for automatic detection of online misogyny written in the Arabic and Levantine dialect\footnote{will be made publicly available on Github.}.\ The proposed dataset consists of 6,550 tweets annotated either as neutral (misogynistic-free) or as one of seven misogyny categories: discredit, dominance, cursing/damning, sexual harassment, stereotyping and objectification, derailing, and threat of violence. The credibility and consistency of the annotations was evaluated through using inter-annotator agreement measures: agreement without chance correction(Cohen's Kappa) and overall Inter-annotator agreement (Krippendorff's alpha).
In addition, we benchmark SOTA approaches on the dataset to evaluate the effectiveness of current approaches for misogyny detection task in the Arabic language.


\section{Related Work}
\label{Sec1}
Since no previous Arabic resources were created for misogyny, and given that misogyny language is considered a type of hate speech, we opted to review the non-Arabic misogyny datasets, and the Arabic abusive and hate speech datasets proposed in the literature. We focus on the characteristics of the presented Arabic annotated resources: data source, data size the tackled toxic categories (for non-misogyny datasets), and annotation/collection strategies.

\subsection{Misogyny Detection in Non-Arabic Languages}
Misogyny detection has been investigated only in a set of languages; precisely in English, Spanish, and Italian. To the best of our knowledge, there are only three datasets in the literature. The work in~\cite{fersini2018ibereval} proposed the first English and Spanish misogyny detection datasets to organize a shared task on automatic detection of misogyny (AMI). The datasets were collected using three different ways: 1) using seed words like \textit{bi**h}, \textit{w**re}, \textit{c*nt}, etc., to collect tweets; 2) monitoring accounts of potential victims (e.g. well known feminist women); 3) downloading tweets from the history of misogynist Twitter accounts, i.e. accounts that explicitly declared hate against women in their Twitter bio (screen name). After collecting tweets, the authors used CrowdFlower platform~\footnote{The platform name changed to \textit{Appen}: \href{https://appen.com}{https://appen.com}} to annotate the tweets. The sizes of the final datasets are 3,977 and 4,138 tweets for English and Spanish, respectively. The authors in~\citep{fersini2018evalita} organized another shared task on AMI, but with focusing only on English and Italian languages. The authors followed the~\cite{fersini2018ibereval} approach to build their datasets. The size of the final datasets is 5,000 tweets for each language.
Both previous works annotated their datasets for two main sub-tasks: Misogyny Identification (i.e. if the tweet is misogynous or not), and Misogynistic Behaviour and Target Classification (i.e. identifying the type of misogyny and recognition of the targets that can be either a specific user or group of women).

\subsection{Arabic Resources for Online Toxicity}
According to~\citep{al2019detection}, the toxic contents on social media can be classified into: abusive, obscene, offensive, violent, adult content, terrorism, and religious hate speech.\ A detailed review of the datasets that covered online toxicity is provided below.

In \cite{Mubarak}, two datasets were proposed: a Twitter dataset of 1,100 dialectal tweets and a dataset of 32K inappropriate comments collected from a popular Arabic news site and annotated as obscene, offensive, or clean.\ The authors in \cite{Alakrota:18}, provided a dataset of 16K Egyptian, Iraqi, and Libyan comments collected from YouTube. The comments were annotated as either offensive, inoffensive, or neutral.\ The inter-annotator agreement for the whole sample was 71\%.\ The religious hate speech detection was investigated in \cite{Albadi} where a multi-dialectal dataset of 6.6K tweets was introduced. The annotation guidelines included an identification of the religious groups targeted by hate speech.\ The calculated inter-annotator agreement for differentiating religious hate speech was 81\% while this value decreased to 55\% when specifying the religious groups targeted by the hate speech.\ Another type of hate speech was tackled in \cite{Ajlan:2018} where the authors presented a Twitter dataset for bullying detection.\ A dataset of 20K multi-dialectal Arabic tweets was collected and annotated manually with bullying and non-bullying labels.\ In this study, no annotation evaluation was provided. More recently, a Twitter dataset L-HSAB of 6K tweets was introduced in \cite{mulki2019} as a benchmark dataset for automatic detection of Arabic Levantine abusive language and hate speech.\ The inter-annotator agreement metric denoted by Krippendorff's alpha ($\alpha$) was 76.5\% and indicated consistent annotations.

\section{Let-Mi Dataset}
\label{Sec2}
Let-Mi can be described as a political dataset as it is composed of tweet replies scraped from the timelines of popular female journalists/reporters during October 17\textsuperscript{th} protests in Lebanon.\ In the following subsections, we provide a qualitative overview of the proposed dataset, while an annotation evaluation is presented in Section \ref{sec5}.

\subsection{Data Collection}
\label{Sec2-1}
The proposed dataset was constructed out of Levantine tweets harvested using Twitter API\footnote{We used python \textit{Tweepy} library \href{http://www.tweepy.org}{http://www.tweepy.org}.}. The collection process relied on scraping tweet replies written at the timelines of several Lebanese female journalists who covered the protests in Lebanon during the period (October 20- November 3, 2019).\ The accounts of the journalists were selected based on their activity on Twitter and the engagement they get from the people\cite{proteststats}.\ As a result, we identified seven journalist accounts as resources of the tweet replies, who represent different national news agencies in Lebanon. Initially, we retrieved 77,856 tweets, and then we manually removed the non-Levantine tweets to cope with the paper's goal, which is to provide a Levantine dataset.\ We also filtered out the non-textual, Arabic-Arabizi mixed tweets, retweets and duplicated instances.\ In addition, based on regular expressions, we spotted many tweets whose content represents a single hashtag or a sequence of hashtags.\ We opted to remove these tweets as they were non-informative and were written just to make a hashtag trending.\ Moreover, to assure that the collected replies are written to target the journalist herself and not a part of side debates among the users within a thread, we removed tweets that mention accounts other than the journalist's.\ Thus, we ended up with 6,603 direct tweet replies.\ Table \ref{tab:Journalists-info} lists the journalist names\footnote{We masked the journalists' names referring to them as J1, J2,... etc.}, their news agencies, and the number of tweet replies collected from the timeline of each.\ In order to prepare the collected tweets for annotation, we normalized them by eliminating Twitter-inherited symbols, digits, and URLs.\ It should be mentioned that as the hashtags encountered within a tweet can indicate a misogynistic content, we removed the hashtag symbol while retaining the hashtag words.

\begin{table}[!htbp]
\centering
\begin{tabular}{lcccl}
\hline
\textbf{Journalist} & \textbf{News Agency} & \textbf{\#Tweets}\\
\hline
J1 & LBC & 4,677 \\
J2 & Freelancer & 1,012\\
J3 & MTV & 351\\ 
J4 & LBC & 179 \\ 
J5 & LBC & 112 \\
J6 & Aljadeed & 171\\ 
J7 & OTV & 101\\\hline
\end{tabular}
\caption{Journalists and their tweet replies size.}\label{tab:Journalists-info}
\end{table}

\subsection{Annotation Guidelines}
\label{Sec2-2}
\begin{table*}
\centering
\scalebox{0.90}{
\begin{tabular}{lcccl}
\hline \textbf{Label} & \textbf{Example}\\ \hline
\scalebox{.8}{None} &\scalebox{.75}{ \textRL{\foreignlanguage{arabic}{لما بكون الأمر واضح وبتاخدي لمكان آخر بيضرب مصداقيتك}}} \\
                              &\scalebox{.75}{When a matter is clear, and you manipulate it, it affects your credibility} \\\hline
\scalebox{.8}{Discredit}&\scalebox{.75}{ \textRL{\foreignlanguage{arabic}{انت خلقتي حمارة وعشتي جحشة ولح تموتي بغلة}}} \\
                                &\scalebox{.75}{You were born as a donkey, and you're living as a colt, and you will die as a mule}     \\\hline
\scalebox{.8}{Derailing}   &\scalebox{.75}{\textRL{\foreignlanguage{arabic}{هيدي معاقه وصار لازم العالم تتمسخر عليها بلكي بتحس عا حاله عميلة الاعلام}}}\\                             
                                &\scalebox{.75}{She is retarded, people must mock her, maybe she realizes what she's done...Media Rat!}\\\hline
\scalebox{.8}{Dominance}   &\scalebox{.75}{\textRL{\foreignlanguage{arabic}{لو منعرفك زلمه منرد عليك}}}  \\                             
                                &\scalebox{.75}{We’d have answered you if you were a man}\\\hline
\scalebox{.8}{Stereotyping \& Objectification}   &\scalebox{.75}{\textRL{\foreignlanguage{arabic}{مش حضرتك لبتقرري هيدا شي انتي فيكي تقرري شو تلبسي وكيف تتمكيجي مش اكتر}}}  \\       
             &\scalebox{.75}{You're not the one who can decide such a thing, you can just decide what to wear and how to put your makeup}\\\hline
\scalebox{.8}{Threat of violence}   &\scalebox{.75}{\textRL{\foreignlanguage{arabic}{والله والله والله العظيم وين بلاقيك بدي اقتلك قواص ذبح حيالله}}}  \\                             
                                               &\scalebox{.75}{I swear by God that wherever I find you I will shoot you or slay you}\\\hline
\scalebox{.8}{Sexual Harassment}   &\scalebox{.75}{\textRL{\foreignlanguage{arabic}{الواطية}}} J1 \scalebox{.75}{\textRL{\foreignlanguage{arabic}{و سوف نغتصبك أيضا و معك   }}} \\                             
                                                   &\scalebox{.75}{We will rape you and rape  degenerate J1 as well}\\\hline
\scalebox{.8}{Damning}   &\scalebox{.75}{\textRL{\foreignlanguage{arabic}{الله يلعنك ويحرقك ويبعتلك سرطان براسك}}}  \\                             
                                &\scalebox{.75}{May God curse and burn you and put cancer in your brain}\\\hline
\end{tabular}
}
\caption{\label{tab:examples} Tweet examples of the annotation labels.}
\end{table*}

The annotation process for Let-Mi dataset requires labeling the tweets as non-misogynistic, i.e. none or as one of seven misogynistic categories: \textit{discredit, derailing, dominance, stereotyping \& objectification, sexual harassment, threat of violence}, and \textit{damning}.\ It should be noted that, besides the standard misogynistic categories that are usually adopted in the misogyny datasets provided for Indo-European languages \cite{fersini2018ibereval,fersini2018evalita,anzovino2018}, we added the ``\textit{damning}" category which represents a novel misogynistic behavior inspired by the Arabic culture.\ It can be described as the Arabic version of cursing where people ask God to make a woman ill, handicapped, dead, hurt, etc.\ Given the difficulty associated with the recognition of misogynistic behaviors \cite{schmidt2017}, they were clearly defined through our designed annotation guidelines.\ This enabled the annotators to have a unified perspective about misogynistic language categories and contributed to improving the inter-annotator agreement scores.\ Based on the definition of misogynistic behaviors in \cite{bailey2016}, we designed the annotation guidelines for Let-Mi dataset such that the eight label categories are identified as follows:

\begin{itemize}
\item Non-Misogynistic (none): tweets are those instances that do not express any hatred, insulting or verbal abuse towards women.
\item Discredit refers to tweets that combine slurring over women with no other larger intention.
\item Derailing: used to describe tweets that indicate a justification of women abuse while rejecting male responsibility with an attempt to disrupt the conversation in order to refocus it.
\item Dominance: tweets are those that express male superiority or preserve male control over women. 
\item Stereotyping \& objectification: used to annotate tweets that promote a widely held but fixed and oversimplified image/idea of women. This label also refers to tweet instances that describe women’s physical appeal and/or provide comparisons to narrow standards.
\item Threat of violence: used to annotate tweets that intimidate women to silence them with an intent to assert power over women through threats of violence physically.
\item Sexual harassment: used for tweets that describe actions such as sexual advances, requests for sexual favors, and sexual nature harassment.
\item Damning: used to annotate tweets that contain prayers to hurt women; most of the prayers are death/illness wishes besides praying God to curse women.
\end{itemize}
Table \ref{tab:examples} lists the relevant examples to each class.

In addition, we provide target annotation for each tweet found to be misogynistic (fall in one of the seven misogynistic categories), therefore we asked the annotators to tag each misogynistic tweet as belonging to one of the following two target categories:
\begin{itemize}
    \item Active (individual): the text includes offensive tweets purposely sent to a specific target (explicit indication of addressee or mention of the journalist name);
    \item Passive (generic): it refers to tweets posted to many potential receivers (e.g. groups of women).
\end{itemize}

\subsection{Annotation Process}
\label{Sec2-3}
The annotation task was assigned to three annotators, one male and two females Levantine native speakers. Besides the previous annotation guidelines, and based on the domain and context of the proposed dataset,we made the annotators aware of specific phrases and terms which look normal/neutral while they indicate toxicity.\ These phrases/terms are either related to the Lebanese culture or developed during the protests in accordance with the incidents.\ For example, ``\textRL{\foreignlanguage{arabic}{بلوطة}}'' (\textit{oak}), which represents a tree type is actually derived from a Lebanese idiom and usually used to describe someone as a liar.\ 
Also, the word ``\textRL{\foreignlanguage{arabic}{سحسوح}}'' (\textit{a slap}) has recently emerged and used during the protest incidents to express an act of violence. 


\begin{table}[!htbp]
\centering
\begin{tabular}{cc}
\hline 
\textbf{Annotation Case} & \textbf{\#Tweets}\\\hline
       Unanimous agreement& 5,529 \\
       Majority agreement (2 out of 3) &1,021 \\
       Conflicts& 53\\
\hline
\end{tabular}
\caption{\label{tab:annotation-stats}Summary of annotation statistics.}
\end{table}

Having all the annotation rules setup, we asked the three annotators to label the 6,603 tweets as either non-misogynistic or one of the seven misogynistic language categories.\ When exploring the annotations obtained for the whole dataset, we faced three cases:
\begin{enumerate}
\item Unanimous agreement: the three annotators annotated a tweet with the same label.\ This was encountered in 5,529 tweets.
\item Majority agreement: two out of three annotators agreed on a label of a tweet.\ This was encountered in 1,021 tweets.
\item Conflicts: each annotator annotated a tweet differently. This case was found in 53 tweets.
\end{enumerate}

After excluding the tweets having three conflicted judgments, the final version of Let-Mi composed of 6,550 tweets.\ A summary of the annotation statistics is presented in Table \ref{tab:annotation-stats}. 

\subsection{Annotation Results}
\label{sec3}
With the annotation process accomplished, and considering the annotation cases in Table \ref{tab:annotation-stats}, the final label of each tweet was determined.\ For tweets falling under the unanimous annotation case, the final labels were directly deduced, while for those falling under the majority annotation case, we selected the label that has been agreed upon by two annotators out of three.\ Consequently, we got 3,388 non-misogynistic tweets and 3,162 misogynistic tweets where the latter were distributed among the seven misogynistic categories.\ A detailed review of the statistics of Let-Mi final version is provided in Table \ref{tab:class-stats} where Voc. denotes the vocabulary size for each class.\ 
\begin{table}[!htbp]
\centering
\scalebox{0.9}{
\begin{tabular}{cccc}
\hline         
\small \bf Label & \small \textbf{\#Tweets}&\small \textbf{\#Words} & \small \textbf{Voc.}\\ \hline
\small None &\small 3,388 &\small 28,610& \small 12,763\\
\small Discredit&\small 2,327&\small 16,587   & \small 6,817   \\
\small Stereotyping \& objectification &\small290 &\small 2,235   &\small 1,426   \\
\small Damning&\small 256 &\small1,479   &\small 868   \\
\small Threat of violence &\small175& \small    1,356   &\small 971   \\
\small Derailing&  \small      59            &  \small   497   & \small 391   \\
\small Dominance&  \small     38           &  \small   292	  &\small 228   \\
\small Sexual harassment & \small      17          & \small   94   &\small 87   \\
\hline
\end{tabular}
}
\caption{\label{tab:class-stats}Tweets distribution across Let-Mi classes.}
\end{table}

To identify the words commonly used within misogynistic contexts, we investigated the lexical distribution of the dataset words across the misogynistic classes.\ Therefore, we subjected Let-Mi to further normalization, where we removed stopwords based on a manually built Levantine stopwords list.\ Later, we identified the five most frequent words and their frequencies in the misogynistic classes as seen in Table \ref{tab:WordFreqs2}, where Dist. denotes the word's distribution in a specific class.
\begin{table}[!htbp]
\centering
\begin{tabular}{cccc}
\hline  \small\bf Dominance                                     &\small\bf Dist.&\small\bf Violence                                               &\small \bf Dist. \\ \hline
\small \textRL{\foreignlanguage{arabic}{السيد}} &\small 3.77\% &\small J1 &\small1.90\%  \\
\small \textit{Al-Sayyid} &&\small \textit{J-name} &  \\
\small\textRL{\foreignlanguage{arabic}{راسك}}&\small 3.42\% &\small\textRL{\foreignlanguage{arabic}{راسك}} &\small1.17\% \\
\small\textit {your head}& &\small\textit{your head}& \\
\small\textRL{\foreignlanguage{arabic}{اشرف}} &\small3.41\%&\small\textRL{\foreignlanguage{arabic}{موتي}}&\small0.66\%\\
\small\textit {more honest})&&\small\textit{be dead} &\\
\small\textRL{\foreignlanguage{arabic}{صرمايتو}}& 1.37\%&\small\textRL{\foreignlanguage{arabic}{الواطية}} &\small0.51\%\\
\small\textit {his slippers}& &\small\textit{degenerate} &\\
\small\textRL{\foreignlanguage{arabic}{داعس}}&\small1.03\% &\small\textRL{\foreignlanguage{arabic}{سحسوح}}&\small0.44\%\\
\small\textit{step on}&&\small\textit{slap}&\\
\hline
\end{tabular}
\caption{\label{tab:WordFreqs2}Distribution of top 5 frequent terms in dominance/threat of violence classes.}
\end{table}

\subsection{Annotation Evaluation}
\label{sec5}
\begin{table*}[!htbp]
\centering
\scalebox{0.98}{
\begin{tabular}{cccc}
\hline 
\small \textbf{Tweet}&\small \textbf{F1} &\small \textbf{F2}&\small \textbf{M1}\\\hline
\small \textRL{\foreignlanguage{arabic}{يلا انزلي انتي جاي عبالك سحسوح مرتب شطة}}& \small violence&\small violence &\small derailing\\
\small \textit{Come on! Try to be at the protests, it seems that you’d like a perfect slap} & & &\\
\small \textRL{\foreignlanguage{arabic}{عقبال القتلة الجاية ههههه}}&\small violence&\small derailing&\small violence\\
\small \textit{Wishing you a next beating hhhh}& & &\\
\small \textRL{\foreignlanguage{arabic}{تروح تنقبر وتلتهي بالبوتوكس وموتوكس وتترك السياسة يحرق روحا}}&\small stereotyping &\small stereotyping&\small damning \\
\small \textit{She should be taking care of her Botox and abandon politics; Damn her soul!} & & &\\
\hline
\end{tabular}
}
\caption{\label{tab:disagreements}Disagreements among annotators.}
\end{table*}
We evaluated the annotations using two inter-annotator agreement measures: Cohen’s kappa \cite{mchugh2012} and  Krippendorff's $\alpha$ \cite{krippendorff2011}.\ For the obtained annotations, we found that pairwise Cohen's Kappa between the two female annotators (F1,F2) was 86.2\%, while for the annotator pairs (male, female), the value decreased to 81.6\% and 80.8\%, respectively.\ Moreover, the calculated Krippendorff's $\alpha$ was 82.9\% which is considered ``good" \cite{krippendorff2011} and indicates the consistency of the annotations.\ Aiming to investigate the disagreement among the annotators, we explored the tweets for which the annotators gave different judgments.\ A sample of the tweets along with the annotations assigned by the annotators is listed in Table \ref{tab:disagreements} where F and M denote female and male, respectively.

As seen in Table \ref{tab:disagreements}, the disagreements spotted among the annotators can be justified by the gender of the annotator where both female annotators had the same judgment for tweets that describe women-related issues (e.g. Botox) besides they were more sensitive to violent threats compared to the male annotator.\ On the other hand, when it comes to sarcastic tweets, regardless of the gender, the annotators provided different class labels due to the normal difference of sense of humor among them.

\section{Experiments and Evaluation}
In this section, we describe our experiments on the Let-Mi data.\ We evaluate the performance of SOTA models on our dataset.\ We design our experiments at three levels (tasks):

\begin{enumerate}
\item Misogyny identification (Binary): tweets contents are classified into  \textit{misogynistic} and \textit{non-misogynistic}.\ This requires merging the seven categories of misogyny into the misogyny class.

\item Categories classification (Multi-class): tweets are classified into categories: \textit{discredit, dominance, damning, derailing, sexual harassment, stereotyping and objectification, and threat of Violence}, or \textit{non-misogynistic}.

\item Target classification (Multi-class): tweets are classified into either \textit{passive, active}, or \textit{non-misogynistic}.
\end{enumerate}

\subsection{Models}
To present a diverse evaluation on the Let-Mi dataset, we test various approaches that use different text representations.\ In the following, we present the used approaches:

\begin{enumerate}
    \item BOW + TF-IDF: word ngrams model with TF-IDF weighting scheme.\ We test several classifiers on a validation part, and we select Naive Bayes classifier\footnote{We tested Logistic Regression (LR), Random Forest (RF), Naive Bayes (NB), and Support Vector Machine (SVM) classifiers}.
    
    \item~\newcite{frenda2018exploration} model: we use one of the SOTA systems on misogyny identification task.\ This model combines character ngrams with several lexicons created by the authors to highlight important cues in the misogynistic tweets.\ For instance, the \textit{Stereotypes} lexicon contains words related to the stereotypes about women, like cooking, taking care of children, etc.\ Since this approach was proposed for English and Spanish languages, we used Google Translation API to translate the lexicons to the Arabic language.\ It is worthy of mentioning that we discard the \textit{Abbreviations} lexicon as it is untranslatable.\ Following the authors' configurations, and we use an ensemble technique (majority voting) to combine the predictions from three different classifiers (NB, SVM, and LR).
    
    \item LSTM: A Long Short-Term Memory (LSTM) neural network that uses Aravec Arabic word embeddings~\citep{soliman2017aravec} to represent texts.\ The output representation of the LSTM is fed to a softmax layer.
    
    \item BERT: is a text representation model that showed leading performance on multiple NLP benchmarks~\citep{devlin2019bert}.\ Since BERT was trained for the English language, we used AraBert~\citep{antoun2020arabert} which is a version trained on Arabic texts.\ We fed the hidden representation of the special [CLS] token that BERT uses to summarize the full input sentence, to a softmax layer.
    
\end{enumerate}

\subsection{Experimental Setup}
Given that Let-Mi dataset is not balanced, especially in the category classification task, we split the dataset into training, validation, and test sets using stratified sampling technique; we take a random 20\% of the tweets from each class for the validation and test sets.\ Discarding the classes' size in the splitting process may affect the minority classes (e.g., sexual harassment).\ For the preprocessing of the text, we remove all special characters, URLs, users mentions, and hashtags to ensure that the evaluation models are not biased to any Twitter-inherited symbol. Regarding the experiments' metrics, we used accuracy and macro precision, recall, and F1 score.

\subsection{Results}
In Table \ref{tab:res_mis_class}, we present the results of the misogyny identification task.\ Considering the F1 metric, the best result of this task is achieved using the BERT model, performing better than the rest of the models.\ We can see that~\citep{frenda2018exploration} model performs slightly better than the BOW and the LSTM models.\ The results show that the LSTM-based model has the lowest performance.

\begin{table}[h]
    \centering
    \scalebox{0.83}{
    \begin{tabular}{ccccc}
    \hline
    \textbf{Model} & \textbf{Acc.} & \textbf{Prec.} & \textbf{Rec.} & \textbf{F1} \\
    \hline
    Majority class  & 0.52 & 0.26 & 0.50 & 0.34 \\ 
    BOW + TF-IDF    & 0.84 & 0.84 & 0.84 & 0.84 \\ 
    \newcite{frenda2018exploration} model & 0.85 & 0.86 & 0.85 & 0.85 \\ 
    LSTM            & 0.82 & 0.82 & 0.82 & 0.82 \\ 
    BERT            & 0.88 & 0.89 & 0.88 & \textbf{0.88} \\ \hline
    \end{tabular}
    }
    \caption{Results of the misogyny identification task.}
    \label{tab:res_mis_class} 
\end{table}

Regarding the category classification task, the results in Table \ref{tab:res_cat_class} show a different scenario.\ The best performing model in misogyny identification task, BERT, performs weaker than BOW and~\citep{frenda2018exploration} models.\ We speculate that this is due to the number of instances for each class.\ Recent studies~\citep{edwards2020go} showed that the performance of the pre-trained models (e.g. BERT) decreases marginally when the number of instances in the training data is less than $\sim$5K, and this was evident for multi-class classification datasets.\ In this task, the total size of the training data is $\sim$4K, and for most of the classes, the maximum number of instances is $\sim$200.

\begin{table}[h]
    \centering
    \scalebox{0.83}{
    \begin{tabular}{ccccc}
    \hline
    \textbf{Model} & \textbf{Acc.} & \textbf{Prec.} & \textbf{Rec.} & \textbf{F1} \\
    \hline
    Majority class  & 0.52 & 0.06 & 0.12 & 0.09 \\
    BOW + TF-IDF    & 0.79 & 0.52 & 0.37 & 0.41 \\
    \newcite{frenda2018exploration} model & 0.81 & 0.62 & 0.38 & \textbf{0.43} \\
    LSTM            & 0.75 & 0.42 & 0.3 & 0.33 \\
    BERT            & 0.81 & 0.44 & 0.33 & 0.35 \\ \hline
    \end{tabular}
    }
    \caption{Results of the categories classification task.}
    \label{tab:res_cat_class} 
\end{table}

For the third task, the results in Table \ref{tab:res_tar_class} show that all the models have a very competitive performance, with a small improvement for the~\citep{frenda2018exploration} model.

\begin{table}[h]
    \centering
    \scalebox{0.83}{
    \begin{tabular}{ccccc}
    \hline
    \textbf{Model} & \textbf{Acc.} & \textbf{Prec.} & \textbf{Rec.} & \textbf{F1} \\
    \hline
    Majority class  & 0.52 & 0.17 & 0.33 & 0.23 \\
    BOW + TF-IDF    & 0.82 & 0.86 & 0.77 & 0.81 \\
    \newcite{frenda2018exploration} model & 0.85 & 0.88 & 0.8 & \textbf{0.83} \\
    LSTM            & 0.82 & 0.82 & 0.81 & 0.82 \\
    BERT            & 0.83 & 0.85 & 0.82 & 0.82 \\ \hline
    \end{tabular}
    }
    \caption{Results of the target classification task.}
    \label{tab:res_tar_class} 
\end{table}

\subsection{Multi-task Learning}
\begin{figure}[h]
\centering
\includegraphics[width=5.5cm]{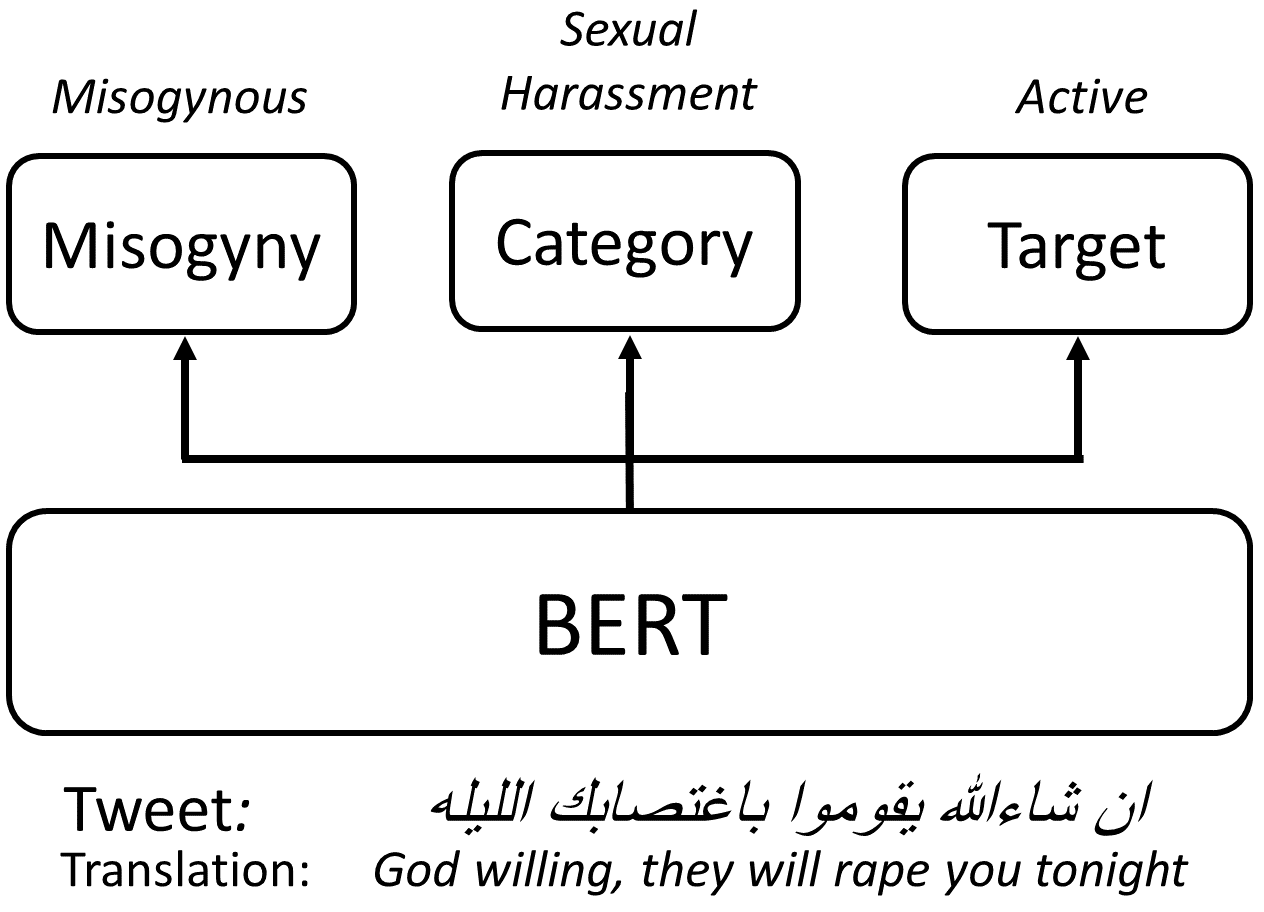}
\caption{BERT model in an MTL configuration.}
\label{fig:MTL}
\end{figure}

In multi-task learning (MTL), a set of relevant tasks (two or more) are involved in the training process of a model to improve the performance on each of them~\citep{caruana1997multitask}.\ MTL enables a model to use cues from various tasks to improve each of them performance, or of a target task~\citep{zhang2019multi}. MTL has been employed in several NLP tasks~\citep{kochkina2018all,majumder2019sentiment,samghabadi2020aggression}.\ In this work, we use our three tasks in an MTL configuration to investigate whether the performance of the model on each of them will improve.\ Therefore, we use the BERT model in our experiment.\ In Figure \ref{fig:MTL} we illustrate the MTL configuration with an example.

The MTL experiment results listed in Table \ref{tab:res_MTL} show that MTL improves the performance of both misogyny and target classification tasks clearly, with an average improvement of \%4 on the F1 metric.\ However, the category classification task's performance does not improve and decreases slightly compared to the BERT performance without MTL (1\% drop).

\begin{table}[h]
    \centering
    \scalebox{0.85}{
    \begin{tabular}{ccccc}
    \hline
    \textbf{Task} & \textbf{Acc.} & \textbf{Prec.} & \textbf{Rec.} & \textbf{F1} \\
    \hline
    Misogyny identification    & 0.9  & 0.9  & 0.9  & 0.9 \\
    Categories classification  & 0.82 & 0.42 & 0.32 & 0.34 \\
    Target classification      & 0.89 & 0.89 & 0.88 & 0.88 \\ \hline
    \end{tabular}
    }
    \caption{Results of the MTL experiment using BERT model.}
    \label{tab:res_MTL} 
\end{table}

\subsection{Error Analysis}
\begin{figure}[h]
\centering
\includegraphics[width=7cm]{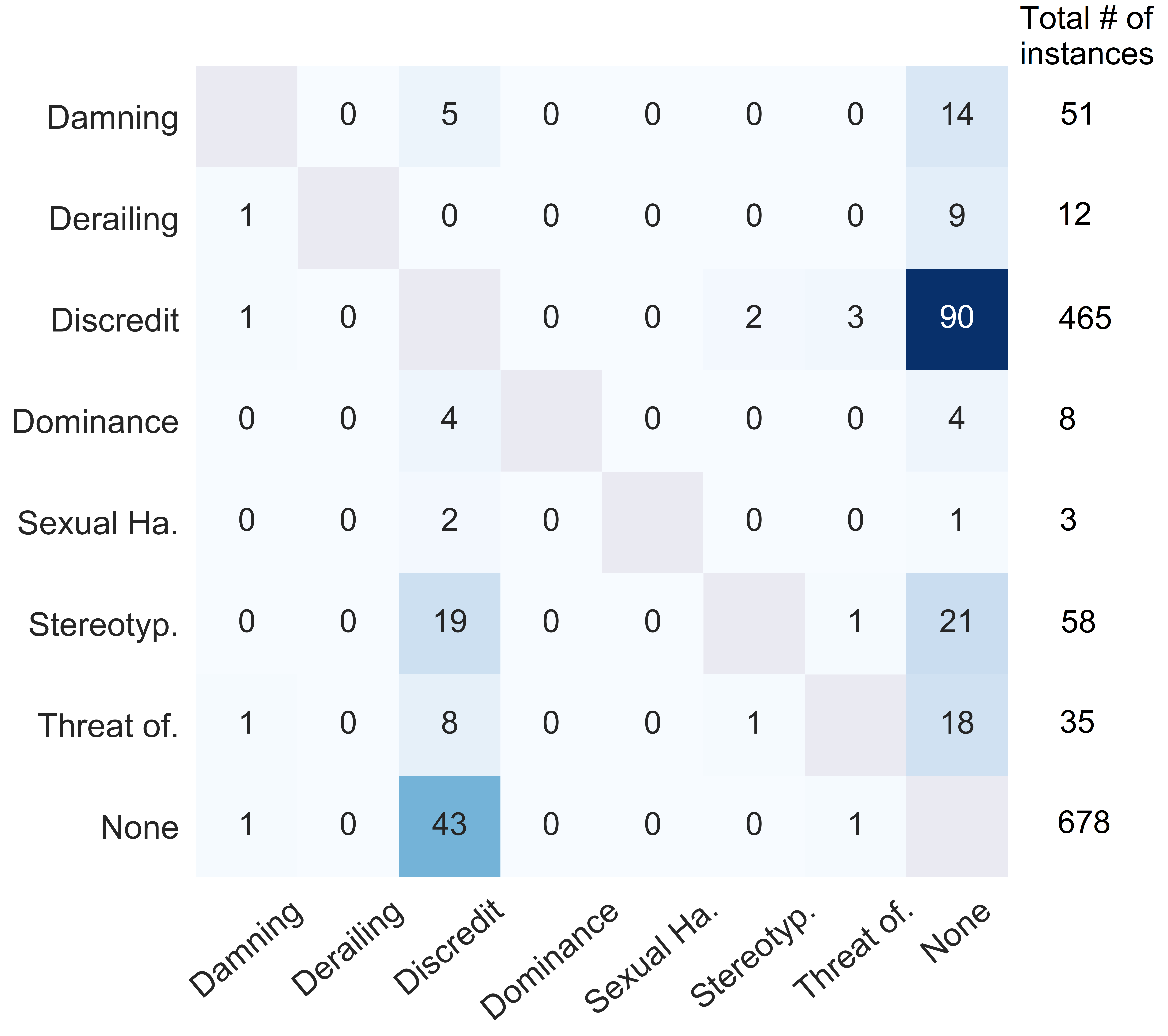}
\caption{Confusion matrix of errors.}
\label{fig:cm}
\end{figure}

Given the previous sections' results, we can notice that the models' performances in the second task are weak compared to the other two tasks.\ Thus, in this section, we investigate the causes of the miss-classified cases in the second task.\ A manual analysis allows us to see which category is the most difficult to be dealt with in the best performing system.\ Figure \ref{fig:cm} reports the confusion matrix of gold labels (y-axis) vs. predicted labels (x-axis).

The manual analysis of the tweets shows that classification errors are due to five main factors:

\begin{enumerate}
    \item \textbf{Weak predictions}: although the tweets have been annotated correctly, the model cannot always detect the misogyny category.\ The cause of this type of errors could be due to the model's generalizability; it could be noted that the model is not able to learn few training cases of some misogynous textual patterns; Besides, this type of errors is evident in the categories having the lowest number of instances in the training data such as \textit{derailing}, \textit{dominance}, and \textit{sexual Harassment}.
    
    \item \textbf{Mixed labels}: few tweets can belong to more than one misogyny category, but the annotators chose the most dominant one.\ In this type of errors, we find that the model sometimes predicts the other possible category.\ For instance, the tweet: ``\textRL{\foreignlanguage{arabic}{انتي مذيعه يا تقبريني خليكي بتلبي عزايم وولائم}}''\footnote{Explanatory translation: are you a journalist! stay at home to cook food for others.} annotated as \textit{Stereotyping \& Objectification} because the author is offending the target women by asking her to stay at home to cook food.\ It can also be considered discredit since the tweet's beginning is discrediting the targeted woman for her performance as a journalist.
    
    \item \textbf{Sarcasm}: many tweets annotated under one of the misogyny categories due to their sarcastic meaning.\ Unfortunately, the model is not able to detect the sarcastic sense.\ For instance, the tweet: ``\textRL{\foreignlanguage{arabic}{كيف صرتي مذيعة؟}}''\footnote{Translation: How you became a journalist?} is sarcastically offending the target woman by meaning that she should not be a journalist.
    
    \item \textbf{Unique damnation phrases}: some tweets use a unique way or a phrase to curse and attack women.\ Similar to other languages, the Arabic language contains many phrases for damnation.\ Apart from the well known phrases, some authors created their own phrases that need some cultural knowledge to be understood.\ An example on these cases is the tweet: ``\textRL{\foreignlanguage{arabic}{ان شاءالله بسقطو زلاعيمك على المسالك البولية}}''\footnote{Literal translation: God willing, your pharynx will fall on the urinary tract.}. This tweet's indirect meaning is that the author prays to God that the journalist will be muted person.
    
    \item \textbf{False misogyny cases}: as we are interested in detecting misogynistic tweets, Let-Mi also contains none-misogynistic tweets to classify both groups.\ Many of the none-misogynistic tweets are neutral tweets (e.g. opinions, questions, news, etc.), but there are also general offensive tweets; not against women.\ Some of these tweets are offences to the misogynist authors.\ The analysis shows that the model sometimes cannot discriminate between offensive tweets that target women and the universal ones.\ An example on this type is the following tweet: 
    ``\textRL{\foreignlanguage{arabic}{\foreignlanguage{english}{1J} اشرف منكو ومن اللي خلفكو انتو اللي واطيين يا عبدت الفاسدين والحرامية}}''\footnote{Literal translation: J1 is more honest than you and your parents, you who are dirty, and slaves of the corrupt and thieves.}.
    
\end{enumerate}


\section{Conclusion and Future Work}
This work proposes Let-Mi, the first misogyny detection dataset for the Levantine dialects of the Arabic language.\ The dataset is annotated manually by a set of Levantine native speakers.\ We present a detailed description of the whole annotation process.\ Additionally, we present an experimental evaluation of several machine learning systems, including SOTA systems.\ Also, we employ an MTL configuration to investigate its effect on the tasks.\ The results show that the performances of the used systems are consistent with SOTA results on the other languages, and MTL improved the performance of the used model on two of the proposed tasks.\ In future work, we plan to create a multi-label version of our dataset for the categories classification task as we found that some tweets can be considered under two or more categories.\ Moreover, since we found many misogynistic tweets that are sarcastic in our analysis, we plan to study the correlation between sarcasm and misogyny tasks.

\bibliography{eacl2021}
\bibliographystyle{acl_natbib}
\end{document}